%% file: manuscript.tex
\documentclass{article}

\usepackage{arxiv}

\usepackage[utf8]{inputenc} 
\usepackage[T1]{fontenc}    
\usepackage{hyperref}       
\usepackage{url}            
\usepackage{booktabs}       
\usepackage{amsfonts}       
\usepackage{nicefrac}       
\usepackage{microtype}      
\usepackage{lipsum}
\usepackage{graphicx}
\graphicspath{ {./images/} }

\usepackage{authblk}

\usepackage{amsmath,amsfonts}
\usepackage{algorithm}
\usepackage{array}
\usepackage[caption=false,font=normalsize,labelfont=sf,textfont=sf]{subfig}
\usepackage{textcomp}
\usepackage{stfloats}
\usepackage{url}
\usepackage{verbatim}
\usepackage{cite}
\usepackage{rotating}
\usepackage{comment}

\usepackage{hyperref}
\usepackage{multirow}

\usepackage{tikz}
\usetikzlibrary{positioning,arrows.meta}
\usepackage{xcolor}
\usepackage{algpseudocode}
\usepackage{colortbl}
\usepackage{booktabs}

\title{FedSLIM: Privacy-Preserving Federated MDL-Based Descriptive Pattern Mining Across Data Silos}

\usepackage{authblk}

\author[1,2]{Samar Samir Khalil}
\author[1]{Noha S. Tawfik}
\author[2,3]{Marco Spruit}

\affil[1]{%
Computer Engineering Department\\
Arab Academy for Science, Technology and Maritime Transport\\
Alexandria 21913, Egypt\\
\texttt{\{samar, noha.abdelsalam\}@aast.edu}
}

\affil[2]{%
Leiden Institute of Advanced Computer Science\\
Leiden University\\
2333 CC Leiden, The Netherlands\\
\texttt{\{s.s.khalil, m.r.spruit\}@liacs.leidenuniv.nl}
}

\affil[3]{%
Department of Public Health \& Primary Care\\
Leiden University Medical Center\\
2333 ZA Leiden, The Netherlands
}

\begin{document}
\maketitle
\begin{abstract}
Federated learning has achieved considerable success for predictive modelling, yet federated descriptive analytics remains largely unexplored. Existing federated pattern mining approaches are predominantly support-based and do not optimise a principled global objective such as Minimum Description Length (MDL). We introduce FedSLIM, the first federated MDL-based framework for descriptive pattern mining. Building on the SLIM principle, FedSLIM enables collaborative optimisation of compact pattern models across distributed databases without sharing raw transactions. We propose two complementary variants that balance privacy, communication, and optimisation fidelity under different deployment assumptions. To evaluate federated MDL mining, we introduce fidelity and discovery-oriented metrics that quantify agreement with a centralised baseline and assess recovery of globally informative patterns. Experiments on multiple real-world datasets under IID and non-IID partitioning show that both variants preserve high-quality compression structure while requiring orders of magnitude less search than the centralised baseline. We further reveal a local-global discovery gap in distributed MDL mining, where globally compressive patterns may be undiscoverable through isolated local optimisation. Both variants recover globally informative patterns absent from all standalone local models, demonstrating the benefits of federated optimisation beyond independent local mining. These results establish federated MDL mining as a practical foundation for privacy-preserving descriptive analytics across distributed data silos. 
\end{abstract}


\section{Introduction}
\label{sec1}
Pattern mining is a fundamental task in data mining, aimed at discovering structured and interpretable patterns from data \cite{fournier2022pattern}. Beyond predictive modelling, it provides descriptive insights that support traceability, hypothesis generation, and domain knowledge discovery. In high-stakes fields like healthcare, finance, and cybersecurity, interpretable summaries are often better than black-box predictive models \cite{rudin2019stop}.

Minimum Description Length (MDL)–based frequent pattern mining methods address a long-standing limitation of traditional frequent pattern mining: pattern explosion \cite{galbrun2022minimum}. Instead of enumerating all patterns above a support threshold, MDL-based approaches select a compact and non-redundant set of patterns that minimises the total description length of the data and the model. By directly optimising compression, MDL-based algorithms produce concise, interpretable summaries that capture the most informative structure in the data. 

However, modern data are rarely centralised. In many real-world applications, data are distributed across multiple institutions, organisations, or devices. For example, clinical records are stored across hospitals, financial transactions across institutions, and user behaviour across edge devices. Direct centralisation of such data is often infeasible or prohibited due to privacy regulations, legal constraints, or data ownership policies. This has led to the emergence of federated learning \cite{mcmahan2017communication}, a paradigm that enables collaborative model training across decentralised data sources without sharing raw data, thereby supporting privacy-preserving learning in sensitive domains.

Despite the rapid development of federated learning techniques, existing work has predominantly focused on predictive models such as neural networks and gradient-based optimisation \cite{khalil2024exploring, zhang2024recent, abreha2022federated}. In contrast, federated descriptive mining, particularly compression-based pattern selection, remains largely unexplored. Existing federated pattern mining methods typically extend frequent itemset or rule mining frameworks by aggregating distributed support counts~\cite{wu2023privacy, mudumba2024mine}. These approaches inherit fundamental limitations: reliance on arbitrary support thresholds, generation of redundant or excessively large pattern sets, and sensitivity to local-versus-global frequency discrepancies in which a pattern may be infrequent locally but frequent globally. More importantly, they do not optimise a principled global objective such as description length.

Federating MDL-based pattern mining introduces several challenges. The MDL objective is inherently global, requiring statistics aggregated across distributed clients to evaluate compression gain consistently. Direct sharing of MDL statistics may also expose sensitive structural information about local datasets, motivating privacy-aware aggregation mechanisms. In addition, under fragmented or non-IID data distributions, globally useful patterns may become locally undiscoverable, causing standalone local optimisation to diverge from the true global compression structure. Finally, communication efficiency becomes a central concern, as iterative codetable optimisation can require repeated exchange of high-dimensional statistics across the federation.

In this work, we introduce, to the best of our knowledge, the first federated framework for MDL-based pattern mining, where multiple clients collaboratively learn a global pattern set that minimises description length without sharing raw data. We instantiate this framework through a federated version of the SLIM algorithm~\cite{smets2012slim}. Our contributions are as follows:

\begin{itemize}
\item We propose \textbf{FedSLIM}, the first federated framework for MDL-based pattern mining, enabling collaborative optimisation of a global codetable across decentralised clients under the MDL principle.

\item We develop two complementary optimisation variants with different privacy and observability trade-offs: \textbf{FedSLIM-SA}, which performs secure globally coordinated aggregation through secure aggregation, and \textbf{FedSLIM-SO}, which enables scalable federated optimisation through selective candidate evaluation.

\item We empirically investigate the \textbf{local--global discovery gap} in federated MDL mining, showing that data partitioning can render globally compressive patterns undiscoverable through isolated local optimisation. We introduce quantitative metrics to measure this phenomenon and evaluate the ability of federated mining to recover the missing global structure.

\item We provide an extensive empirical evaluation across multiple datasets and federated partitioning schemes, demonstrating that federated MDL optimisation can preserve the dominant compression structure while approximating centralised compression quality.
\end{itemize}

The rest of this paper is organised as follows. Section~\ref{sec2} summarises prior research on MDL for itemset mining and federated association rules mining. Section~\ref{sec3} explains the proposed model with its variants. The experimental work and its results are discussed in sections~\ref{sec4} and~\ref{sec5}, followed by the discussion and conclusion in sections~\ref{sec6} and~\ref{sec7}. 

\section{Related Work}
\label{sec2}
Since federated MDL-based pattern mining remains unexplored, we position this work at the intersection of MDL-based pattern mining and federated pattern mining. We first review MDL approaches and justify the choice of SLIM, then examine federated methods, showing that they are largely support-driven and do not optimise global compression objectives.

\subsection{Minimum Description Length for Pattern Mining}
The Minimum Description Length principle provides an objective, statistically motivated criterion for pattern selection: good models are those that compress the data well when the model itself is included in the encoding~\cite{galbrun2022minimum}. 

In transactional pattern mining, KRIMP~\cite{vreeken2011krimp} introduced the first MDL-based framework, selecting a compressive subset from a pre-mined collection of frequent itemsets generated under a minimum support threshold. SLIM~\cite{smets2012slim} extends this idea by eliminating the dependence on a fixed candidate pool and instead directly refining the codetable through iterative candidate generation and greedy selection. This makes SLIM less reliant on exhaustive frequent itemset mining and generally more effective at discovering compact, non-redundant pattern sets. Subsequent work improved scalability and diversity of search~\cite{sampson2014widened}, explored alternative combinatorial formulations such as data packing~\cite{tatti2008finding}, and investigated evolutionary and meta-heuristic miners for compression-based pattern sets~\cite{nawaz2024genetic,nawaz2025grimp}. 

Broader MDL pattern mining also spans settings beyond standard categorical market-basket transactions, including compressing sequential patterns~\cite{lam2014mining}, real-valued intervals~\cite{witteveen2014realkrimp}, tree-structured data~\cite{hess2014shrimp}, numerical pattern collections~\cite{makhalova2022mint}, MDL analysis of database similarity~\cite{budhathoki2015difference}, and anomaly detection perspectives rooted in compression~\cite{akoglu2012fast}.

We instantiate our federated framework in the itemset–transaction setting using SLIM. While KRIMP introduced the MDL-based codetable framework, its dependence on a pre-mined candidate pool makes it a less practical foundation for extension. SLIM's direct codetable refinement through iterative candidate generation eliminates this stage and achieves stronger compression, making it the more effective starting point. Building on federated learning, which initially used simple averaging as a straightforward aggregation method, we selected SLIM as the foundation for federated MDL since its optimisation depends on using globally aggregated usage statistics.

\subsection{Federated association rules and frequent pattern mining}

When data are partitioned across multiple parties, frequent pattern and association rule mining must rely on protocols that expose only the minimum information necessary to estimate supports and evaluate rule quality. Early secure distributed mining addressed horizontal partitioning through cryptographic and disclosure-limited designs~\cite{tassa2013secure}.

Subsequent federated formulations have taken several directions. Molina et al.~\cite{molina2021federated} replaced raw frequency with an interestingness measure in a coordinator–node architecture, enabling discovery of rare but highly correlated patterns in EHRs. FedFPM~\cite{wang2022fedfpm} introduced an interactive query–response protocol in which the server generates Apriori-based candidates, clients return noisy binary responses under local differential privacy, and Hoeffding's inequality guides adaptive pruning. Wu et al.~\cite{wu2023privacy} coupled federated learning with a pre-large concept to reduce repeated database scans in IIoT settings, securing local outputs via third-party encryption before central aggregation. Saini et al.~\cite{saini2024privacy} secured federated market basket analysis through partial homomorphic encryption, enabling the server to perform additive aggregation entirely in the encrypted domain.

A persistent limitation across these frameworks is their dependence on Apriori-based enumeration, which demands multiple dataset scans and incurs substantial overhead. FedFIM~\cite{chen2023privacy} addressed this by applying the NegFIN algorithm to reconstructed noisy datasets, significantly reducing training time. Mudumba et al.~\cite{mudumba2024mine} proposed a mine-first paradigm, extracting rules independently per source before integration, improving memory efficiency over conventional integrate-first approaches. Panos et al.~\cite{panos2025comparative} and Fernandez et al.~\cite{fernandez2024designing} extended secure federated mining along complementary directions, comparing SMC-based and homomorphic encryption-based paradigms, and generalising Tassa's~\cite{tassa2013secure} framework to fuzzy association rule mining via level-wise $\alpha$-cut decomposition. 

A critical observation is that many methods described as federated are, more precisely, decentralised/distributed mining approaches: they perform local mining independently and aggregate patterns only post hoc, rather than collaboratively optimising a shared global objective. This distinction is core to our work, where the goal is to minimise description length over all distributed data jointly, not merely to consolidate local outputs.

\section{Methodology}\label{sec3}
This section presents the proposed federated MDL framework. We first review the centralised SLIM algorithm, then present our federated contribution, detailing the key design decisions and how they address the challenges of the distributed setting.

\subsection{Centralised SLIM}
Given a transactional database $\mathcal{D}$ over item universe $\mathcal{I}$, SLIM maintains a codetable $CT$ consisting of itemsets paired with their assigned codes. It is initialised with the standard code table $ST$, which contains only singleton itemsets, ensuring every transaction can be encoded. Each transaction $t \in \mathcal{D}$ is encoded via a cover procedure under the standard cover order. The order prioritises itemsets by decreasing cardinality, then decreasing usage, then lexicographically. The output of $\mathrm{cover}(t, CT)$ is a set of non-overlapping itemsets assigned to $t$, with any remaining items covered by singleton entries.

The \emph{usage} of each itemset $X \in CT$ is defined as the number of transactions whose cover contains $X$. Usage counts determine code lengths under Shannon-optimal coding,
\begin{equation}
    \label{eq:code_length}
    L(c_X) = -\log_2 \frac{\mathrm{usage}(X)}{\sum_{Y \in CT}\mathrm{usage}(Y)},
\end{equation}
so that frequently used itemsets receive shorter codes. The overall quality of $CT$ is measured by the MDL objective
\begin{equation}
    L(CT, \mathcal{D}) = L(CT \mid \mathcal{D}) + L(\mathcal{D} \mid CT),
\end{equation}
which balances the cost of describing the codetable against the cost of encoding the data with it.

Starting from $CT = ST$, SLIM iteratively proposes candidate itemsets by combining pairs of itemsets existing in $CT$, accepting a candidate only if its insertion reduces the total description length $L(CT, \mathcal{D})$. After each accepted insertion, coverages, usages, and code lengths are recalculated. A pruning step then removes any existing itemsets whose deletion further improves compression. The process repeats until no candidate yields improvement, producing a compact codetable that summarises the dominant co-occurrence structure in $\mathcal{D}$.

\subsection{Federated SLIM}

\input{figures/pseudo}

Adapting SLIM to the federated setting requires addressing three  interdependent constraints: global usage must be reconstructed from distributed local covers without sharing raw transactions; candidates must be generated and consolidated across clients with heterogeneous local codetables; and client-level usage statistics must be protected from server-side inference. The proposed framework addresses all three through a server-orchestrated, synchronous protocol with three phases: 
(1)~\emph{initialization}, in which clients register with the server and report their protected $ST$ usage statistics;
(2)~\emph{iterative refinement}, in which the server broadcasts a request for candidates, clients generate and report candidate statistics, the server selects the best candidate, evaluates it via a second round of client querying, and either accepts or rejects it; and 
(3)~\emph{convergence}, when no client can generate a candidate that improves global compression. Algorithm~\ref{alg:fedslim} presents the complete protocol, and Figure~\ref{fig:example} provides a step-by-step illustration of a single iteration on a toy two-client database.

\input{figures/example}

All client-server communication is protected using mutual TLS (mTLS)~\cite{rfc8446} under a public key infrastructure (PKI) with a server-managed certificate authority (CA)~\cite{rfc5280}, providing authenticated and encrypted transport between the server and participating clients. Messages are transmitted as length-prefixed, zlib-compressed~\cite{rfc1950} serialised payloads.

The two variants, \textbf{FedSLIM-SA} and \textbf{FedSLIM-SO}, share this protocol structure but differ in how clients report usage statistics to the server and in the degree of inter-client transparency the protocol affords, as detailed in the following subsections. To further limit server-side disclosure, clients adopt a shared item encoding scheme prior to federation. Each item $it \in \mathcal{I}$ is mapped to an opaque integer identifier, so that all itemsets communicated to the server are expressed in terms of these identifiers rather than their semantic labels. Under this scheme, the server operates over anonymised codetables and cannot interpret the content of any itemset it aggregates. Sharing the encoding with the server is optional and may be appropriate when the application context permits it, when the server is a trusted coordinator rather than an untrusted aggregator.

\noindent\textbf{FedSLIM-SA (Secure Aggregation / Server-Oblivious):}
\indent FedSLIM-SA augments federated SLIM with cryptographic secure aggregation so that the server observes only exact aggregate usage statistics and never individual client contributions. It learns solely arithmetic sums; it cannot recover any client's usage values or distinguish a client with zero usage for an itemset from one that never held it locally.

The central challenge is that standard secure aggregation~\cite{bonawitz2017secagg} requires all clients to submit vectors of identical length and positional alignment so that pairwise masks cancel entry-wise upon summation. In gradient-based federated learning this holds naturally, as all clients share the same model architecture. In federated SLIM it does not, as each client maintains a local codetable $CT_i$ over a client-specific subset of itemsets, making usage vectors structurally heterogeneous. Applying masks over misaligned vectors would cause cancellation at wrong positions, corrupting the aggregate.

We address this by separating structure from values across two rounds. In Round~1, clients transmit only the structural skeleton of their local state, consisting of either the ordered itemsets in $CT_i$ or the locally generated candidate tuples $(XY, X, Y)$, without any numerical usage values. The server aggregates these, constructs a deterministic global index $I_{\mathrm{global}}$ ordered by the standard cover order, and broadcasts it to all clients. 

In Round~2, each client $C_i$ constructs a padded vector $\mathbf{v}_i$ aligned with this index, where the $k$-th entry corresponds to the $k$-th itemset in $I_{\mathrm{global}}$:
\begin{equation}
    [\mathbf{v}_i]_k =
    \begin{cases}
        \mathrm{usage}_i\!\left(I_{\mathrm{global}}[k]\right), & \text{if } I_{\mathrm{global}}[k] \in CT_i,\\
        0, & \text{otherwise.}
    \end{cases}
\end{equation}
Each client then applies standard pairwise masking and submits the result, allowing the server to recover only the aggregated usage values. For candidate generation, clients instead submit a padded $|I_{\mathrm{cand}}| \times 3$ matrix containing the usages of $XY$, $X$, and $Y$ after insertion.

This two-round protocol is applied uniformly across all aggregation steps in the federated SLIM loop, including usage collection, candidate generation, and exact MDL evaluation. During evaluation, each candidate $XY$ is broadcast to all $n$ clients. Clients that generated it compute the corresponding updated usage statistics, while the remaining clients return their unchanged usage series, padded and masked to the common evaluation index $I_{\mathrm{eval}} = I_{\mathrm{global}} \cup \{XY\}$. Because all responses share the same masked and padded structure, the server cannot distinguish participating from non-participating clients. Padding also contributes to privacy: a zero entry is indistinguishable from the absence of the corresponding itemset in a client’s codetable, partially concealing the contribution structure in addition to the values themselves.

At the cryptographic layer, pairwise secrets are established once at registration via elliptic-curve Diffie--Hellman (ECDH), from which each client derives $n-1$ pairwise seeds and expands them into per-round mask vectors that cancel exactly upon summation at the server. Mask independence across protocol stages is enforced by incorporating a round identifier and a per-aggregation sub-round counter into the key derivation, preventing reuse. Under the semi-honest model (honest-but-curious) and standard hardness assumptions on the underlying elliptic curve, each masked vector is computationally indistinguishable from uniform noise to any observer lacking the pairwise secrets, and the server learns nothing beyond aggregate sums. The contribution of FedSLIM-SA lies not in these cryptographic primitives, which are standard, but in their orchestration to support exact secure aggregation over structurally heterogeneous codetables, a setting that standard secure aggregation protocols do not directly address.

\noindent\textbf{FedSLIM-SO (Server-Observable / Client-Isolated):}
\indent FedSLIM-SO adopts a direct reporting approach in which each client transmits its exact local usage statistics to the server without any cryptographic protection on the values themselves. The server therefore observes the precise usage counts contributed by each client and can identify which clients generated which candidates. Privacy is enforced along a different axis. As in both variants, all itemsets are exchanged exclusively through their opaque integer encodings so the server aggregates and reasons over anonymised pattern identifiers and cannot interpret the semantic content of any itemset it processes unless the encoding is explicitly shared.

\begin{figure*}[t]
\centering
\resizebox{\textwidth}{!}{%
    \input{figures/flow}
}
\caption{Communication and execution sequence of \textsc{FedSLIM-SA} and \textsc{FedSLIM-SO}}
\label{fig:fedslim}
\end{figure*}
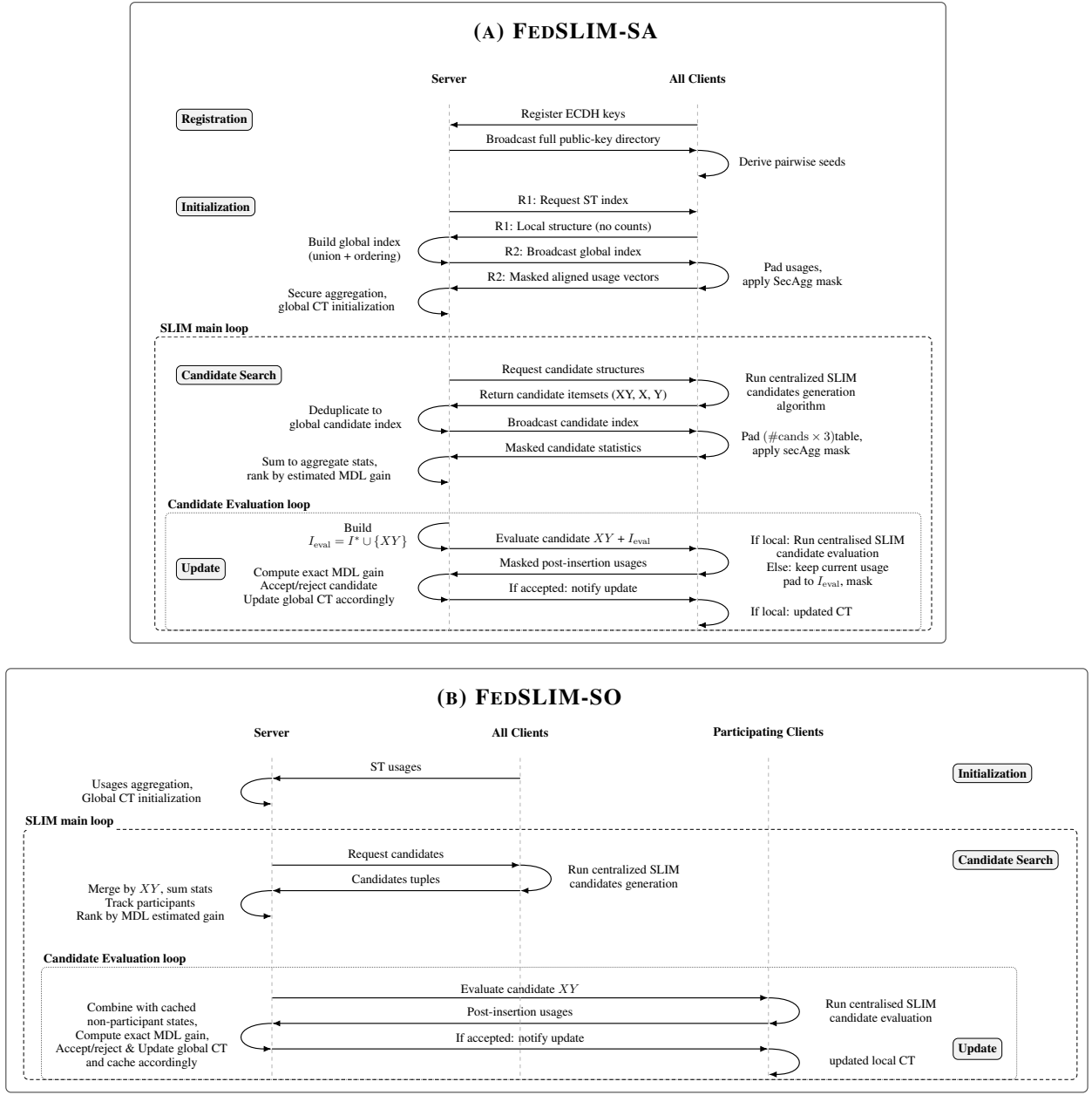

Figure \ref{fig:fedslim}, shows the communication workflow of the two variants. FedSLIM-SO differs from FedSLIM-SA in two architecturally significant ways. First, because clients submit exact usage values directly without requiring global index alignment or pairwise masking, the two-round structure of FedSLIM-SA is unnecessary. Each client transmits its local usage series in a single round per protocol phase, and the server aggregates by direct summation, resulting in substantially lower per-round communication overhead. Second, candidate evaluation is selective rather than universal. During each iteration, the server broadcasts a \textsc{request\_candidates} message to all clients and records which clients generated each candidate. When evaluating a specific candidate $XY$, only participating clients, those with local evidence for $XY$, are queried via \textsc{evaluate\_candidate}. Non-participating clients, whose codetables are unaffected by $XY$, are not contacted; instead, the server retrieves their most recently received usage vectors from a server-side cache and folds these directly into the evaluation aggregate. The cache is updated only when a client's codetable changes, that is, when it participates in an accepted candidate, and remains valid between updates since a non-participating client's usage values are stable across any candidate it did not generate. This selective querying reduces per-evaluation communication from $O(n)$ to $O(m)$, where $m \ll n$ is the number of participating clients, a saving that compounds significantly across the large number of candidates evaluated per iteration.

The inter-client isolation of FedSLIM-SO is a direct consequence of its selective evaluation design. Since each candidate pattern is communicated only between the server and its generating clients, no client is ever exposed to itemsets discovered or maintained by other participants. Clients therefore operate independently throughout the protocol, contributing local statistics to the server without any visibility into the broader federation. Only the server maintains a complete view of the global codetable, making FedSLIM-SO particularly suitable for deployments in which the server is the intended consumer of the analysis while clients must remain mutually opaque.

Importantly, clients are not merely passive contributors but also benefit from the federation itself. By aggregating usage evidence across participants, the server can recover globally informative patterns that appear uninformative on every individual client because their supporting evidence is distributed across shards. Once accepted into the global codetable, these patterns are redistributed to relevant clients, enabling compression improvements that standalone local mining would not discover. The federation therefore overcomes the fragmentation effect by recovering globally useful structure that remains locally undiscoverable under isolated optimisation.

\section{Experimental Work}\label{sec4}

\subsection{Datasets}
The proposed models are evaluated on eight benchmark datasets from the SLIM implementation~\cite{slim_impl} and the Frequent Itemset Mining Implementations (FIMI) repository~\cite{FIMI}, spanning three orders of magnitude in dataset size. Table~\ref{tab:datasets} summarises their key statistics. The datasets range from the small Ionosphere collection to the large-scale Accidents corpus. Most datasets exhibit fixed-length transactions, reflecting their origin as transformed tabular data. Accidents is a notable exception, with variable-length transactions ranging from 18 to 51 items. Owing to this variability, together with its large scale ($\sim$340K transactions and 468 items), it serves as the primary dataset in our experiments.

\begin{table}[ht]
\centering
\caption{Datasets statistics.}
\label{tab:datasets}
\setlength{\tabcolsep}{5pt}
\begin{tabular}{lrrr}
\toprule
\textbf{Dataset} & \textbf{Trans\#} & \textbf{Items\#} & \textbf{Trans Length} \\
\midrule
Ionosphere       &    351   &   157 & Fixed: 35\\
Chess (k-k)      &  3,196   &    75 & Fixed: 37\\
Mushroom         &  8,124   &   119 & Fixed: 23\\
Pendigits        & 10,992   &    86 & Fixed: 17\\
Letter Recog.    & 20,000   &   102 & Fixed: 17\\
Adult            & 48,842   &    97 & Variable: 12-15\\
Connect-4        & 67,557   &   129 & Fixed: 43\\
Accidents        &340,183   &   468 & Variable: 18-51\\
\bottomrule
\end{tabular}
\end{table}

\subsection{Experimental Setup}
\label{subsec:setup}
The proposed algorithms are implemented in Python using standard scientific and systems libraries. The full implementation is publicly available on GitHub\footnote{\url{https://github.com/Samar67/FedSLIM}.} We evaluate our methods against the original SLIM results reported in~\cite{smets2012slim}, denoted as \emph{Centralised}. To ensure a fair and controlled comparison, we additionally run the centralized \emph{scikit-mine} implementation of SLIM~\cite{sk_mine} under identical hardware and software conditions as our proposed federated models. We adopt the \emph{scikit-mine} implementation specifically because it provides full programmatic access to the resulting codetable, enabling us to compute evaluation metrics beyond those reported in the original paper. Following prior work, all experiments are subject to a maximum runtime of 24 hours~\cite{smets2012slim}. Experiments were conducted on the ALICE high-performance computing cluster~\cite{alice}, using compute nodes from the cpu-zen4 partition based on modern AMD Zen4 processors. Each job was allocated a single task with 32 CPU cores and 32 GB of RAM, ensuring consistent and reproducible computational conditions across all experiments.

\subsection{Evaluation Metrics}

Prior work on federated pattern mining has largely focused on efficiency-oriented measures such as execution time, communication cost, and the number of extracted patterns, while MDL-based pattern mining studies typically evaluate compression performance through description length and codetable size. Such metrics provide only a partial view of federated MDL-based mining, as they do not assess whether the discovered patterns faithfully reproduce the centralised solution or whether federated learning can recover globally informative patterns that are undiscoverable from isolated local data. To provide a more comprehensive evaluation, we assess the proposed federated algorithms across five dimensions: compression quality, model characteristics, communication efficiency, fidelity to the centralised baseline, and local-versus-global discovery.

To assess compression quality, we use the compression ratio, defined as
\begin{equation}
  L(\%) = \frac{L(\mathcal{D} \mid CT)}{L(\mathcal{D} \mid ST)}
  \times 100,
  \label{eq:lperc}
\end{equation}
where $L(\mathcal{D} \mid CT)$ is the encoded length of database $\mathcal{D}$ under the learned codetable $CT$, and $L(\mathcal{D} \mid ST)$ is the length under the standard codetable $ST$. Lower values indicate better compression and more informative itemset selection. We also report the number of non-singleton codetable elements $|CT|$, which reflects the algorithm's ability to identify meaningful higher-order structures, and the number of candidate itemsets evaluated $|F|$, which serves as a proxy for computational cost. For communication efficiency, we record the total number of server-client interaction rounds $R$ and the total number of bytes transmitted required for convergence $BT$.

To measure how closely the federated codetable reproduces the centralised solution, we compare $A$ with $B$, where $A$ denotes the set of non-singleton itemsets contained in the centralised \emph{scikit-mine} codetable and $B$ the set of non-singleton itemsets contained in the federated codetable. Recall and precision are defined as
\begin{equation}
  \mathrm{Recall} = \frac{|A \cap B|}{|A|}, \qquad
  \mathrm{Precision} = \frac{|A \cap B|}{|B|},
  \label{eq:rec_prec}
\end{equation}
which measure, respectively, the fraction of centralised itemsets recovered by the federated model and the fraction of federated itemsets that are consistent with the centralised solution. Using these quantities, we compute the F1-score,
\begin{equation}
  F_1 = 2 \cdot
  \frac{\mathrm{Precision} \cdot \mathrm{Recall}}
  {\mathrm{Precision} + \mathrm{Recall}},
  \label{eq:f1}
\end{equation}
which provides a balanced measure of structural agreement between the centralised and federated codetables.

To evaluate whether the federated approach preserves the most important compressive patterns, we additionally report Weighted Recall@$k$ (WR@$k$), which measures the fraction of total centralised usage contributed by the top-$k$ centralised itemsets that are recovered by the federated model. Finally, we compute Spearman's rank correlation coefficient ($\rho$) over the usage counts of matched itemsets to evaluate whether the federated codetable preserves the relative importance ordering of patterns discovered by the centralised model.

To quantify the limitations of standalone local MDL-based mining under fragmented data distributions, we evaluate whether the proposed federated variants can recover globally compressive patterns that are not discoverable through isolated local optimisation. Let $\mathcal{I}(CT_i)$ denote the set of non-singleton itemsets discovered by the standalone local codetable of client $i$.

We first define the union of all locally discovered itemsets as $ \textstyle\bigcup_i CT_i$, which represents all patterns recoverable through independent local SLIM executions. Using this union, we define the \emph{discovery gap} as
\begin{equation}
  G = A \setminus \textstyle\bigcup_i CT_i,
  \label{eq:gap}
\end{equation}
where $G$ contains the globally selected itemsets that are absent from every local codetable simultaneously. These itemsets correspond to patterns that provide sufficient compression benefit at the global level yet fail to produce enough local compression gain to be selected by any client in isolation. To evaluate the ability of the federated framework to recover these globally informative yet locally undiscoverable patterns, we measure the \emph{gap recovery rate},
\begin{equation}
  GRR = \frac{|G \cap B|}{|G|},
  \label{eq:recovery}
\end{equation}
This metric quantifies the fraction of missing global patterns successfully recovered by the federated approach despite their absence from all standalone local codetables.

\section{Experimental Evaluation and Results}
\label{sec5}
We evaluate the proposed algorithms under a comprehensive set of controlled experimental scenarios designed to isolate the key challenges of federated MDL-based pattern mining. We first assess robustness under ideal IID partitioning across multiple datasets, followed by scalability with increasing federation size, robustness to client-level data imbalance, and resilience under non-IID heterogeneity in both support-skew and combined support-and-quantity-skew settings. Finally, we examine the local-global discovery gap, a structural fragmentation phenomenon in which globally informative itemsets become locally undiscoverable because their supporting evidence is distributed across clients.

Collectively, these experiments characterise the behaviour of FedSLIM across a broad range of federated data conditions, enabling analysis of its robustness to fragmentation, scalability, imbalance, and distribution heterogeneity.

\subsection{IID Robustness Across Datasets.}
To assess robustness under ideal federated conditions, we evaluate both variants under IID equal partitioning across 10 clients on eight datasets spanning different scales and codetable complexities. Each client receives an equal-sized random partition of the dataset, providing a controlled setting for evaluating how closely the federated algorithms approximate the centralised solution in the absence of intentionally introduced distribution shift. 

\begin{table*}[t]
\centering
\small
\caption{
IID robustness across 10 equally sized clients.
$L\%$ denotes compression ratio, $|CT|$ codetable size, $|\mathcal{F}|$ evaluated candidates, $R$ communication rounds, and $BT$ transmitted bytes. F1 measures codetable overlap with the centralised baseline, WR@50 recovery of the highest-impact patterns, and $\rho$ Spearman rank correlation over matched usages.
}

\label{tab:iid}
\begin{tabular}{llrrrrrrrrr}
\toprule
Dataset & Algorithm & L\% & $|\mathit{CT}|$ & $|\mathcal{F}|$ & R & BT & F1 & WR@50 & $\rho$ \\
\midrule
 \multirow{3}{*}{Ionosphere} & Centralised & 49.70 & 240 & 294k & --- & --- & --- & --- & --- \\
 & FedSLIM-SA & 50.35 & 255 & 679 & 294 & 3.6 GB & 0.818 & 0.985 & 0.907 \\
 & FedSLIM-SO & 50.35 & 258 & 749 & 308 & 542.1 MB & 0.813 & 0.985 & 0.910 \\
 \midrule
\multirow{3}{*}{Chess(k-k)} & Centralised & 14.70 & 292 & 20k & --- & --- & --- & --- & --- \\
 & FedSLIM-SA & 15.15 & 250 & 546 & 373 & 2.22 GB & 0.883 & 0.933 & 0.978 \\
 & FedSLIM-SO & 15.38 & 248 & 553 & 369 & 608.3 MB & 0.871 & 0.905 & 0.976 \\
\midrule
 \multirow{3}{*}{Mushroom} & Centralised & 18.50 & 340 & 16k & --- & --- & --- & --- & --- \\
 & FedSLIM-SA & 18.66 & 490 & 1352 & 718 & 12.91 GB & 0.387 & 0.674 & 0.752 \\
 & FedSLIM-SO & 18.79 & 462 & 1195 & 682 & 3.29 GB & 0.406 & 0.674 & 0.724 \\
 \midrule
 \multirow{3}{*}{Pen digits} & Centralised & 39.40 & 1347 & 394k & --- & --- & --- & --- & --- \\
 & FedSLIM-SA & 45.37 & 786 & 815 & 796 & 36.27 GB & 0.353 & 0.934 & 0.572 \\
 & FedSLIM-SO & 43.96 & 1359 & 2434 & 1415 & 53.1 GB & 0.379 & 0.934 & 0.652 \\
 \midrule
 \multirow{3}{*}{Letter Recognition} & Centralised & 33.40 & 1599 & 521k & --- & --- & --- & --- & --- \\
 & FedSLIM-SA & 38.20 & 737 & 766 & 761 & 29.74 GB & 0.304 & 0.811 & 0.622 \\
 & FedSLIM-SO & 36.10 & 1643 & 2833 & 1792 & 94.23 GB & 0.359 & 0.826 & 0.750 \\
 \midrule
  \multirow{3}{*}{Adult} & Centralised & 22.80 & 1201 & 199k & --- & --- & --- & --- & --- \\
 & FedSLIM-SA & 23.95 & 831 & 895 & 863 & 33.22 GB & 0.673 & 0.949 & 0.786 \\
 & FedSLIM-SO & 23.80 & 1037 & 1524 & 1095 & 19.53 GB & 0.645 & 0.949 & 0.772 \\
 \midrule
 \multirow{3}{*}{Connect-4} & Centralised & 12.30 & 1670 & 297k & --- & --- & --- & --- & --- \\
 & FedSLIM-SA & 14.18 & 786 & 954 & 921 & 45.67 GB & 0.442 & 0.624 & 0.676 \\
 & FedSLIM-SO & 13.66 & 1501 & 1969 & 1718 & 140.65 GB & 0.476 & 0.624 & 0.760 \\
 \midrule
  \multirow{3}{*}{Accidents} & Centralised & 31.10 & 2018 & 21k & --- & --- & --- & --- & --- \\
 & FedSLIM-SA & 39.87 & 364 & 368 & 367 & 20.52 GB & 0.469 & 0.759 & 0.661 \\
 & FedSLIM-SO & 33.99 & 1429 & 1489 & 1448 & 111.34 GB & 0.545 & 0.772 & 0.807 \\
\bottomrule
\end{tabular}
\end{table*}

Table~\ref{tab:iid} shows that both federated variants remain competitive with the centralised baseline while evaluating one to two orders of magnitude fewer candidates. Across all datasets, the federated search typically explores fewer than $3\times10^3$ candidates, compared with up to $5.2\times10^5$ centrally, indicating that strong compression can be recovered with substantially reduced search effort.

On smaller datasets such as Ionosphere and Chess (k-k), both variants closely match the centralised codetable, with F1 above $0.81$, WR@50 above $0.90$, and near-perfect rank correlation. As dataset size and codetable complexity increase, compression fidelity and codetable overlap begin to diverge. On Pendigits, Letter Recognition, Connect-4, and Accidents, both methods maintain competitive compression despite substantially lower F1 scores, indicating that multiple codetable configurations can achieve similar MDL quality in large transactional spaces.

This distinction is most visible in the importance-aware metrics. Even when F1 falls below $0.40$, WR@50 remains comparatively high, reaching $0.934$ on Pendigits and $0.949$ on Adult. The federated codetables therefore continue to preserve the dominant compression structure despite reduced set-level agreement with the centralised solution. These results suggest that MDL quality depends primarily on recovering a relatively small subset of high-impact patterns rather than reproducing the exact codetable.

The two variants exhibit distinct optimisation characteristics. FedSLIM-SA consistently evaluates fewer candidates taking fewer rounds but incurs substantial communication overhead due to dense full-participation aggregation. FedSLIM-SO exchanges more rounds and transmitted bytes, yet preserves stronger fidelity on larger datasets by enabling broader candidate exploration. This effect is most pronounced on Accidents, where FedSLIM-SO nearly matches the centralised compression ratio (33.99\% vs.\ 31.10\%) while substantially outperforming FedSLIM-SA in F1, WR@50, and rank correlation.

\subsection{IID Scalability with Number of Clients.}
To evaluate scalability, we vary the number of participating clients while keeping the data distribution IID. Using the Accidents dataset, we construct federations with $n \in \{2, 4, 8, 16, 32, 64, 128\}$ clients, where data is evenly partitioned among clients in each configuration. This experiment examines how the proposed models behave as the federation scales while isolating the effect of reduced local data availability. 

\begin{table*}[t]
\centering
\small
\caption{
Scalability under increasing federation size on Accidents.
IID partitioning is maintained while varying the number of clients.
}
\label{tab:scalability}
\begin{tabular}{llrrrrrrrrr}
\toprule
Clients\# & Algorithm & L\% & $|\mathit{CT}|$ & $|\mathcal{F}|$ & R & BT & F1 & WR@50 & $\rho$ \\
\midrule
2 & FedSLIM-SA & 35.77 & 849 & 877 & 861 & 21.92 GB & 0.702 & 0.792 & 0.818 \\
& FedSLIM-SO & 33.62 & 1771 & 1867 & 1794 & 45.46 GB & 0.495 & 0.792 & 0.750 \\
\midrule
4 & FedSLIM-SA & 37.76 & 536 & 546 & 544 & 16.9 GB & 0.612 & 0.792 & 0.724 \\
& FedSLIM-SO & 33.81 & 1611 & 1699 & 1634 & 66.43 GB & 0.524 & 0.792 & 0.765 \\
\midrule
8 & FedSLIM-SA & 38.74 & 447 & 453 & 453 & 23.75 GB & 0.557 & 0.779 & 0.669 \\
& FedSLIM-SO & 34.08 & 1449 & 1525 & 1471 & 94.21 GB & 0.559 & 0.792 & 0.773 \\
\midrule
16 & FedSLIM-SA & 42.98 & 238 & 241 & 240 & 16.37 GB & 0.352 & 0.759 & 0.533 \\
 & FedSLIM-SO & 34.56 & 1158 & 1193 & 1174 & 97.56 GB & 0.611 & 0.772 & 0.840 \\
\midrule
32 & FedSLIM-SA & 47.64 & 134 & 134 & 134 & 14.31 GB & 0.220 & 0.525 & 0.502 \\
 & FedSLIM-SO & 35.97 & 777 & 797 & 786 & 66.42 GB & 0.679 & 0.772 & 0.852 \\
\midrule
64 & FedSLIM-SA & 53.20 & 78 & 78 & 78 & 14.34 GB & 0.130 & 0.261 & 0.506 \\
& FedSLIM-SO & 37.77 & 529 & 539 & 534 & 49.76 GB & 0.590 & 0.772 & 0.745 \\
\midrule
128 & FedSLIM-SA & 71.54 & 16 & 16 & 16 & 4.93 GB & 0.031 & 0.020 & 0.432 \\
& FedSLIM-SO & 38.75 & 445 & 451 & 449 & 62.36 GB & 0.535 & 0.759 &  0.698 \\
\midrule
\multicolumn{5}{l}{\textit{Centralised: $L\%$ = 31.10}} \\
\bottomrule
\end{tabular}
\end{table*}

Table~\ref{tab:scalability} reveals a clear divergence between the two variants as federation size increases. FedSLIM-SA degrades rapidly beyond $n=16$, with compression ratio increasing from 35.77\% at $n=2$ to 71.54\% at $n=128$, while F1 decreases from 0.702 to 0.031. In contrast, FedSLIM-SO remains comparatively stable, maintaining compression close to the centralised baseline together with substantially stronger fidelity metrics. At $n=32$, for example, FedSLIM-SO achieves 35.97\% compression with F1 of 0.679, whereas FedSLIM-SA degrades to 47.64\% with F1 of only 0.220.

The primary limitation is communication cost under the fixed 24-hour execution budget. FedSLIM-SA relies on dense global aggregation for every candidate evaluation, causing communication to dominate optimisation as the number of clients increases. This sharply restricts candidate exploration: at $n=128$, FedSLIM-SA evaluates only 16 candidates and produces a codetable containing 16 non-singleton itemsets. The degradation therefore reflects constrained search depth rather than instability of the MDL objective itself.

FedSLIM-SO scales more favourably because candidate evaluation remains localised, enabling substantially broader exploration within the same time budget. Consequently, WR@50 remains near $0.75$ even at $n=128$, indicating that the dominant compression structure can still be recovered despite increasingly fragmented local evidence.

\subsection{Uneven IID (Data Imbalance) Robustness.}
To isolate the effect of client-level data imbalance independently of distribution shift, we construct an uneven IID setting on the Accidents dataset. Transactions are distributed across 10 clients using a Dirichlet allocation with $\alpha=0.5$, producing substantially different local dataset sizes (Table~\ref{tab:uneven_iid_partition}) while preserving the global data distribution. This setting simulates realistic federated environments in which participants contribute unequal amounts of data.

\begin{table*}[ht]
\centering
\caption{Uneven IID data partitioning across clients.}
\label{tab:uneven_iid_partition}
\setlength{\tabcolsep}{5pt}
\renewcommand{\arraystretch}{1.05}
\begin{tabular}{l rrrrrrrrrr}
\toprule
\textbf{Client} & $C_1$ & $C_2$ & $C_3$ & $C_4$ & $C_5$ & $C_6$ & $C_7$ & $C_8$ & $C_9$ & $C_{10}$ \\
\midrule
\textbf{Data Size} & 91850 & 61230 & 47630 & 34020 & 27210 & 20410 & 17010 & 13610 & 13610 & 13610 \\
\bottomrule
\end{tabular}
\end{table*}

Table~\ref{tab:uneven_iid_results} shows that unequal shard sizes alone have limited impact on compression quality. FedSLIM-SA achieves 41.00\% compression and FedSLIM-SO 34.39\%, remaining close to their balanced IID counterparts. Fidelity metrics exhibit similarly minor variation, indicating that aggregated usage statistics naturally accommodate heterogeneous client contributions without requiring explicit weighting.

Communication behaviour follows the same trend observed under balanced IID partitioning. FedSLIM-SA exchanges fewer total bytes but explores a substantially smaller candidate space, whereas FedSLIM-SO incurs a higher communication cost in exchange for stronger codetable fidelity. Interestingly, both methods reduce transmitted bytes relative to the balanced setting, suggesting that larger shards concentrate statistical evidence and accelerate candidate acceptance. Overall, partition-size imbalance appears substantially less harmful than federation scale or fragmentation of local evidence.

\begin{table*}[t]
\centering
\small
\caption{
Robustness to uneven IID partitioning on Accidents (10 clients).
}
\label{tab:uneven_iid_results}
\begin{tabular}{lrrrrrrrrr}
\toprule
Algorithm & L\% & $|\mathit{CT}|$ & $|\mathcal{F}|$ & R & BT & F1 & WR@50 & $\rho$ \\
\midrule
FedSLIM-SA & 41.00 & 310 & 313 & 314 & 15.65 GB & 0.437 & 0.779 & 0.579 \\
FedSLIM-SO & 34.39 & 1302 & 1361 & 1320 & 85.04 GB & 0.596 & 0.792 & 0.794 \\
\midrule
\multicolumn{5}{l}{\textit{Centralised: $L\%$ = 31.10}} \\
\bottomrule
\end{tabular}
\end{table*}

\subsection{Non-IID Robustness under Support and Quantity Skew.}
To evaluate robustness under statistical heterogeneity, we construct two non-IID settings on the Accidents dataset. The first introduces \emph{support skew}, where clients exhibit different dominant patterns while sharing the same global item space. Dominant itemsets are selected from the centralised \emph{scikit-mine} codetable based on usage and filtered to ensure structural diversity. Transactions are then assigned using a coverage-based rule, where each transaction is allocated to the client whose dominant itemsets are most strongly represented within it. This produces client-specific pattern biases while preserving the overall item vocabulary.

The second setting extends this construction with \emph{quantity skew}, producing clients that differ in both transaction volume and local pattern distributions. Specifically, after assigning transactions according to dominant-pattern coverage, client capacities are determined using a Dirichlet allocation with $\alpha=0.5$. This introduces substantial imbalance in local dataset sizes, causing some clients to receive large transaction shards while others receive only a small fraction of the global data. Consequently, the resulting federation exhibits heterogeneity in both the prevalence of local patterns and the amount of data available at each participant.

To further control the strength of the induced heterogeneity, both settings are evaluated using dominant itemsets of minimum size 3 and 4. Larger dominant itemsets generate more specific local co-occurrence structures, increasing divergence between clients and strengthening the non-IID character of the federation. Table~\ref{tab:non_iid_partitioning} summarises the resulting transaction allocation across clients. The first two columns correspond to support skew only, where client sizes emerge from the coverage-based assignment procedure. The final column corresponds to the combined support--quantity skew setting, where the additional Dirichlet allocation ($\alpha=0.5$) intentionally creates highly imbalanced client shard sizes, ranging from fewer than one hundred transactions to more than 140,000 transactions.

\begin{table}[ht]
\centering
\caption{
Client data allocation under support skew and combined support--quantity skew. In the latter setting, client shard sizes are additionally controlled by a Dirichlet allocation ($\alpha=0.5$).
}
\label{tab:non_iid_partitioning}
\setlength{\tabcolsep}{4pt}
\renewcommand{\arraystretch}{1.05}
\begin{tabular}{c cc c}
\toprule
& \multicolumn{2}{c}{\textbf{Support Skew}} \\
\cmidrule(lr){2-3}
\textbf{Client\#} &
  3-itemsets & 4-itemsets &
  \textbf{Support \& Quantity Skew}\\
\midrule
$C_1$ & 134169 & 150027 & 18282\\ 
$C_2$ & 64235 & 41533 & 85868 \\ 
$C_3$ & 21657 & 38356 & 3171  \\ 
$C_4$ & 21751 & 29016 & 439 \\ 
$C_5$ & 23953 & 15789 & 48956  \\
$C_6$ & 19456 & 13744 & 55  \\ 
$C_7$ & 14336 & 17489 & 142735 \\ 
$C_8$ & 18975 & 10538 & 4307  \\ 
$C_9$ & 10165 & 10391 & 12060 \\ 
$C_{10}$ & 11486 & 13300 & 24310 \\
\bottomrule
\end{tabular}
\end{table}

\begin{table*}[t]
\centering
\small
\caption{
Robustness under support skew and combined support-quantity skew on Accidents.
}
\label{tab:non_iid_results}
\begin{tabular}{cclrrrrrrrr}
\toprule
Skew Type & Min Dom & Algorithm & L\% & $|\mathit{CT}|$ & $|\mathcal{F}|$ & R & BT & F1 & WR@50 & $\rho$ \\
\midrule
\multirow{4}{*}{Support}  & \multirow{2}{*}{3} & FedSLIM-SA & 40.65 & 333 & 341 & 336 & 17.55 GB & 0.394 & 0.673 & 0.639 \\
 &  & FedSLIM-SO &  34.49 & 1410 & 1474 & 1424 & 104.21 GB & 0.471 & 0.651 & 0.809 \\
\cmidrule(lr){2-11}
 & \multirow{2}{*}{4} & FedSLIM-SA & 39.77 & 402 & 412 & 405 & 24.25 GB & 0.438 & 0.487 & 0.617  \\
 &  & FedSLIM-SO &  35.01 & 1388 & 1468 & 1400 & 100.44 GB & 0.451 & 0.471 & 0.760\\
\midrule
\multirow{4}{*}{Support \&}  & \multirow{2}{*}{3} & FedSLIM-SA  & 40.53 & 334 & 338 & 336 & 17.24 GB & 0.378 & 0.616 & 0.639 \\
  &  & FedSLIM-SO  & 34.13 & 1496 & 1585 & 1511 & 99.45 GB & 0.455 & 0.616 & 0.804 \\
\cmidrule(lr){2-11}
 Quantity & \multirow{2}{*}{4} & FedSLIM-SA  & 41.85 & 278 & 281 & 280 & 12.96 GB & 0.333 & 0.653 & 0.582\\
 &  & FedSLIM-SO  & 34.13 & 1498 & 1576 & 1516 & 99.38 GB & 0.459 & 0.636 & 0.796 \\
\midrule
\multicolumn{5}{l}{\textit{Centralised: $L\%$ = 31.10}} \\
\bottomrule
\end{tabular}
\end{table*}

Table~\ref{tab:non_iid_results} evaluates robustness under support skew and combined support-quantity skew. Overall, both variants remain relatively stable despite substantial client heterogeneity. FedSLIM-SA achieves compression ratios between 39.77\% and 41.85\%, while FedSLIM-SO remains closer to the centralised baseline at 34.13\%--35.01\%.

The primary effect of non-IID partitioning appears in pattern fidelity rather than compression quality. FedSLIM-SO consistently preserves stronger rank agreement with the centralised codetable ($\rho=0.760$--$0.809$) together with substantially larger codetables, reflecting broader exploration of globally useful patterns under heterogeneous local distributions. FedSLIM-SA produces smaller codetables and lower fidelity metrics, particularly under stronger heterogeneity, due to the exploration constraints imposed by dense global aggregation within the fixed execution budget.

Increasing the minimum dominant itemset size from 3 to 4 strengthens client-specific co-occurrence structure and reduces WR@50 for both variants, particularly under support skew. Nevertheless, the overall degradation remains moderate, indicating that the dominant compression structure can still be recovered despite substantial distributional divergence across clients.

\subsection{The Local-Global Discovery Gap.}

A fundamental challenge in distributed pattern mining is fragmentation: an itemset may improve compression globally while appearing uninformative on every individual client because its supporting evidence is distributed across shards. For example, in a federation of hospitals, the combination \{smoking history, chronic cough, weight loss\} may appear only a handful of times at each site and therefore fail to be selected by local mining algorithms. However, when evidence from all hospitals is aggregated, the same pattern may occur frequently enough to constitute an important global descriptor of a patient subgroup. In support-based mining, this corresponds to the classical case where an itemset is globally frequent but locally infrequent everywhere. In FedSLIM, the same effect emerges at the level of MDL optimisation and codetable membership, making it an inherent consequence of partitioning rather than a support-threshold artefact.

Under IID partitioning across $n$ clients, the local compression benefit of inserting a candidate itemset $X$ decreases roughly in proportion to shard size, while the codetable cost of representing $X$ remains approximately unchanged. This yields the approximation
\begin{equation}
\Delta L_i(X)
\;\approx\;
\frac{1}{n}\,\Delta L(X)
\;-\;
\frac{n-1}{n}\cdot 2L(c_X),
\label{eq:local_mdl_gain}
\end{equation}
where $\Delta L(X)$ denotes the global compression improvement obtained by inserting $X$, and $L(c_X)$ is its code length under the current codetable. The first term captures dilution of supporting evidence across clients, whereas the second reflects the persistent local model overhead of encoding the pattern. As the number of clients increases, the compression benefit shrinks while the model cost remains relatively stable, causing $\Delta L_i(X)$ to become negative on every client even when $\Delta L(X) > 0$ globally. Consequently, an itemset may be globally useful yet locally undiscoverable on every shard. 

We evaluate the local-global discovery gap on Accidents, Adult, and Chess (k-k), representing large-, medium-, and small-scale transactional settings, respectively. Table~\ref{tab:final_gap_metrics} shows that the gap emerges naturally under standard IID federation: across all datasets, many itemsets selected by the centralised codetable are absent from every local codetable, demonstrating that globally useful MDL patterns can become locally undiscoverable even without distribution shift.

\begin{table}[t]
\centering
\scriptsize
\setlength{\tabcolsep}{3pt}
\caption{Recovery of globally useful but locally undiscoverable itemsets under IID federation.}
\label{tab:final_gap_metrics}
\begin{tabular}{lll rrrr}
\toprule
Dataset & Algorithm & $|G|$ & $|G\!\cap\! B|$ & GRR & WR@50 & WR@100\\ 
\midrule
\multirow{2}{*}{Chess(k-k)} & FedSLIM-SA &  \multirow{2}{*}{184} & 164 & 0.891 & 0.933 & 0.932 \\
 & FedSLIM-SO &  & 158 & 0.859 & 0.905 & 0.911\\
\midrule
\multirow{2}{*}{Adult}  & FedSLIM-SA &  \multirow{2}{*}{843} & 383 & 0.454 & 0.949 & 0.916\\
 & FedSLIM-SO &  & 421 & 0.499 & 0.949 & 0.916\\
\midrule
\multirow{2}{*}{Accidents}  & FedSLIM-SA & \multirow{2}{*}{39} & 6 & 0.154 & 0.759 & 0.737\\
 & FedSLIM-SO & & 14 & 0.359 & 0.773 & 0.752\\
\bottomrule
\end{tabular}
\end{table}

The severity of the discovery gap varies substantially across datasets. Chess (k-k) exhibits the strongest recovery, with both variants recovering more than $85\%$ of globally useful patterns that are absent from every local codetable. This is consistent with its compact codetable and relatively concentrated pattern structure, which limit fragmentation under IID partitioning. Adult presents a substantially larger gap ($|G| = 843$), yet both variants still recover nearly half of the missing patterns while maintaining WR@50 and WR@100 above $0.91$. This indicates that many unrecovered itemsets belong to the low-impact tail of the codetable and contribute little to the dominant compression structure. Accidents is the most challenging setting, where recovery rates decrease noticeably, particularly for FedSLIM-SA. Nevertheless, both variants still preserve strong agreement on the highest-ranked patterns despite recovering only a subset of the globally useful itemsets.

These results highlight an important distinction: exact recovery of every globally useful pattern is not necessary for preserving MDL quality. What matters is recovering the high-impact itemsets that dominate description length reduction, and these are consistently retained through global usage aggregation. Overall, the experiment validates the central premise of FedSLIM: federated MDL optimisation can recover globally informative structure that is invisible to standalone local mining, even under standard IID partitioning.

\section{Discussion}
\label{sec6}
The experiments reveal a consistent trade-off between optimisation fidelity, communication cost, and observability. FedSLIM-SA performs dense globally coordinated optimisation through secure aggregation, limiting server-side visibility into client contributions while operating entirely over encoded item representations. The server, therefore, cannot interpret candidates' semantic meaning unless the encoding is explicitly shared. In exchange for these stronger privacy guarantees, FedSLIM-SA explores a substantially smaller candidate space under fixed execution budgets, making it more suitable for small- to medium-scale collaborative environments where preserving global structure is more important than exact codetable recovery. Examples include multi-institution healthcare consortia, cross-bank fraud analysis, or international data-sharing partnerships operating under data sovereignty constraints, where persistent exposure of per-client statistics carries legal or ethical risk. 

FedSLIM-SO adopts a more selective strategy that enables substantially broader exploration and stronger compression fidelity, particularly under large federations and heterogeneous data distributions. Here, the server observes which client generated each candidate together with its associated usage statistics, while clients remain completely isolated from one another and only access their own local exploration process. This setting is well suited to federated analytics scenarios in which the server is the intended recipient of the analysis, such as enterprise monitoring, multi-site healthcare analytics, or cross-organisational security analysis. It is also appropriate for environments where raw data cannot legally or operationally leave the client side due to regulatory constraints. In such settings, clients still benefit from recovering globally informative patterns that would remain locally undiscoverable under standalone mining, enabling knowledge sharing without exposing the underlying data.

Across all experiments, both variants consistently preserve the dominant compression structure even when codetable overlap with the centralised baseline decreases considerably. This suggests that federated MDL mining depends primarily on recovering high-impact patterns rather than reproducing the exact global codetable. At the same time, the results expose several practical limitations. Communication cost remains substantial for high-dimensional datasets, particularly under dense aggregation, while the fixed synchronous execution budget increasingly constrains exploration as federation size grows. 

To assess whether the remaining gap to the centralised baseline was primarily caused by the 24-hour execution budget, we repeated the non-converged IID experiments using a 72-hour budget. Although the extended runs explored additional candidates, the resulting improvements in compression quality were generally modest relative to the increase in communication cost. This suggests that the 24-hour budget is not the primary factor limiting fidelity to the centralised solution. Detailed results are reported in Table S1 of the supplementary material.

An important consideration is how the proposed methods would behave on substantially larger datasets. Although our experiments are limited to the Accidents dataset, the underlying computational characteristics provide some insight into expected scalability. If the number of transactions were increased by two orders of magnitude while maintaining a similar item vocabulary and pattern structure, the overall runtime would be expected to grow approximately linearly. This is because the dominant operation during both local optimisation and global evaluation is the cover function, whose cost scales linearly with the number of transactions. Under such conditions, larger datasets would primarily increase computation time and communication volume, but would not fundamentally alter the optimisation process.

In contrast, growth in the number of distinct items is likely to present a more significant scalability challenge. As item dimensionality increases, code tables tend to become larger and more diverse, expanding the candidate search space explored during optimisation. Candidate generation and evaluation must then consider a substantially greater number of possible itemset combinations, increasing both computational and memory requirements at the client side and enlarging the set of candidates exchanged during federated coordination. Consequently, the practical scalability limits of FedSLIM are expected to be driven more by the complexity of the item space than by the number of transactions alone.

\section{Conclusion}
\label{sec7}

This work introduced FedSLIM, the first federated MDL-based framework for pattern mining. Building on the SLIM principle, FedSLIM enables collaborative codetable optimisation across distributed clients without raw data sharing, realised through two complementary variants: FedSLIM-SA, which performs globally coordinated aggregation under stronger privacy constraints, and FedSLIM-SO, which enables broader candidate exploration through selective optimisation and stricter client isolation.

Experiments across IID and non-IID partitioning schemes, varying federation scales, and heterogeneous data distributions consistently showed that both variants recover high-quality compression structure while evaluating orders of magnitude fewer candidates than the centralised baseline. Crucially, exact codetable recovery proved unnecessary for strong MDL performance; compression quality depends primarily on identifying a small subset of high-impact itemsets. The study also identified a fundamental local-global discovery gap, that is, globally informative itemsets can become locally undiscoverable under partitioning due to fragmented supporting evidence, even under IID federation. Both variants successfully recover a substantial portion of these patterns through global aggregation, demonstrating that federated MDL mining can reconstruct structure that standalone local mining cannot observe.

The two variants together expose a practical spectrum of trade-offs between optimisation fidelity, communication cost, and observability, supporting deployment under different trust and regulatory assumptions. Key open challenges include the substantial communication overhead on high-dimensional data and the optimisation constraints imposed by synchronous execution at scale. Future work will investigate communication-efficient aggregation strategies, asynchronous optimisation, and formal privacy guarantees under adversarial threat models.

\section*{Acknowledgments}
This work was performed using the compute resources from the Academic Leiden Interdisciplinary Cluster Environment (ALICE) provided by Leiden University.

\bibliographystyle{unsrt} 
\bibliography{biblio}
\end{document}

%% file: figures/pseudo.tex
\begin{algorithm}[t]
\caption{\textsc{FedSLIM}: Federated MDL Codetable Optimisation}
\label{alg:fedslim}
\small
\begin{algorithmic}[1]
\State \textbf{Input:} Clients $\mathcal{C}=\{1,\ldots,n\}$ with local databases $\{D_i\}$
\State \textbf{Output:}  Global codetable $CT$ and description length $L$

\ForAll{$i\in\mathcal{C}$}
    \State $(ST_i,u_i)\gets \textsc{InitSLIM}(D_i)$
\EndFor

\State $CT\gets \textsc{AggregateUsage}(\{ST_i,u_i\})$
\State $L\gets \mathrm{MDL}(CT)$

\Repeat

    \ForAll{$i\in\mathcal{C}$ \textbf{in parallel}}
        \State $\Gamma_i\gets \textsc{GenerateCandidates}(CT_i)$
        \State Compute local candidate statistics
    \EndFor

    \State $A\gets \textsc{AggregateCandidates}(\{\Gamma_i\})$
    \State $\Gamma\gets \textsc{EstimateGain}(A,CT)$

    \State $\mathit{accepted}\gets \mathrm{false}$

    \ForAll{$XY\in\Gamma$ (descending estimated gain)}

        \State $CT^{*}\gets \textsc{EvaluateCandidate}(XY)$

        \If{$\mathrm{MDL}(CT^{*}) < L$}
            \State $CT\gets CT^{*}$
            \State $L\gets \mathrm{MDL}(CT)$
            \State Broadcast acceptance of $XY$
            \State Update participating local codetables
            \State $\mathit{accepted}\gets \mathrm{true}$
            \State \textbf{break}
        \EndIf

    \EndFor

\Until{$\neg\mathit{accepted}$}

\State \Return $CT,L$
\end{algorithmic}
\end{algorithm}

%% file: figures/example.tex
\definecolor{sacolor}{RGB}{0,100,180}       
\definecolor{bkcolor}{RGB}{170,50,0}        
\definecolor{colAB}{RGB}{0,120,50}          
\definecolor{colSng}{RGB}{90,40,140}        
\definecolor{cellOn}{RGB}{210,228,248}      
\definecolor{cellAcc}{RGB}{208,242,210}     

\newcommand{\covAB}[1]{\textcolor{colAB}{\textbf{[}\textit{#1}\textbf{]}}}
\newcommand{\covSg}[1]{\textcolor{colSng}{[\textit{#1}]}}

\newcommand{\cY}{\cellcolor{cellOn}$\bullet$}   
\newcommand{\cN}{\phantom{$\bullet$}}            

%

\begin{figure*}[t]
\centering
\setlength{\tabcolsep}{3pt}
\footnotesize


\noindent\textbf{Step~1: Local data processing and usage reporting}

\smallskip

\begin{minipage}[t]{0.36\linewidth}
\centering
\textit{Client $C_1$ \normalfont($trans\#=5$)}

\smallskip

\begin{tabular}{c|ccccc}
\toprule
      & $a$ & $b$ & $c$ & $d$ & $e$ \\
\midrule
$t_1$ & \cY & \cY & \cY & \cN & \cN \\
$t_2$ & \cY & \cY & \cY & \cN & \cN \\
$t_3$ & \cY & \cY & \cY & \cN & \cN \\
$t_4$ & \cY & \cN & \cN & \cN & \cY \\
$t_5$ & \cY & \cN & \cN & \cN & \cY \\
\bottomrule
\end{tabular}%
\quad
\begin{tabular}{lc}
\toprule
Item & $u_{C_1}$ \\
\midrule
$\{a\}$ & 5 \\
$\{b\}$ & 3 \\
$\{c\}$ & 3 \\
$\{e\}$ & 2 \\
\bottomrule
\end{tabular}
\end{minipage}%
\hfill
\begin{minipage}[t]{0.20\linewidth}
\centering\vspace{12pt}
\textit{Each client sends}\\
\textit{usage counts $u_{C_k}$}\\
\textit{to server}
\end{minipage}%
\hfill
\begin{minipage}[t]{0.36\linewidth}
\centering
\textit{Client $C_2$ \normalfont($trans\#=5$)}

\smallskip

\begin{tabular}{c|ccccc}
\toprule
         & $a$ & $b$ & $c$ & $d$ & $e$ \\
\midrule
$t_6$    & \cY & \cY & \cN & \cY & \cN \\
$t_7$    & \cY & \cY & \cN & \cY & \cN \\
$t_8$    & \cY & \cY & \cN & \cY & \cN \\
$t_9$    & \cY & \cY & \cN & \cY & \cN \\
$t_{10}$ & \cN & \cN & \cY & \cN & \cY \\
\bottomrule
\end{tabular}%
\quad
\begin{tabular}{lc}
\toprule
Item & $u_{C_2}$ \\
\midrule
$\{a\}$ & 4 \\
$\{b\}$ & 4 \\
$\{d\}$ & 4 \\
$\{c\}$ & 1 \\
$\{e\}$ & 1 \\
\bottomrule
\end{tabular}
\end{minipage}

\smallskip
\noindent\rule{\linewidth}{0.4pt}\smallskip


\noindent\textbf{Step~2: Server aggregates usages — global codetable $CT_0$ and MDL baseline}

\smallskip

\begin{minipage}[t]{0.38\linewidth}
\centering
\begin{tabular}{lccc}
\toprule
Itemset  & $U{=}u_{C_1}{+}u_{C_2}$ & $L(c_X)$ \\
\midrule
$\{a\}$  & $5+4=9$  & 1.585 \\
$\{b\}$  & $3+4=7$  & 1.948 \\
$\{c\}$  & $3+1=4$  & 2.755 \\
$\{d\}$  & $0+4=4$  & 2.755 \\
$\{e\}$  & $2+1=3$  & 3.170 \\
\bottomrule
\end{tabular}
\end{minipage}%
\hfill
\begin{minipage}[c]{0.58\linewidth}
\raggedright
$L(c_X)=-\log_2(U_X/\textstyle\sum U)$,\enspace $\sum U=27$\\[4pt]
$L(D\mid CT_0)=9{\cdot}1.585+7{\cdot}1.948+4{\cdot}2.755+4{\cdot}2.755+3{\cdot}3.170=59.4$\,bits\\[2pt]
$L(CT_0\mid D)=2\times(1.585+1.948+2.755+2.755+3.170)=24.4$\,bits\\[2pt]
$\boldsymbol{L_0 = 59.4+24.4 = 83.9}$\,\textbf{bits}
\end{minipage}

\smallskip
\noindent\rule{\linewidth}{0.4pt}\smallskip


\noindent\textbf{Step~3: Local candidate generation, server aggregation, gain estimation and ranking}

\smallskip

\begin{minipage}[t]{0.26\linewidth}
\centering
\textit{$C_1$ valid local candidates}

\smallskip
\begin{tabular}{lc}
\toprule
Candidate & Local $u$ \\
\midrule
$\{a,b\}$ & 3 \\
$\{a,c\}$ & 3 \\
$\{a,e\}$ & 2 \\
$\{b,c\}$ & 3 \\
\bottomrule
\end{tabular}
\end{minipage}%
\hfill
\begin{minipage}[t]{0.26\linewidth}
\centering
\textit{$C_2$ valid local candidates}

\smallskip
\begin{tabular}{lc}
\toprule
Candidate & Local $u$ \\
\midrule
$\{a,b\}$ & 4 \\
$\{a,d\}$ & 4 \\
$\{b,d\}$ & 4 \\
$\{c,e\}$ & 1 \\
\bottomrule
\end{tabular}
\end{minipage}%
\hfill
\begin{minipage}[t]{0.42\linewidth}
\centering
\textit{Server: aggregated global usage + estimated gain $\widehat{\Delta L}$}

\smallskip
\begin{tabular}{lcc}
\toprule
Candidate & Global $U$ & $\widehat{\Delta L}$ (bits) \\
\midrule
\rowcolor{cellAcc}$\{a,b\}$ & 7 & $-13.8$ \\
$\{b,d\}$ & 4 & $-7.4$ \\
$\{a,d\}$ & 4 & $-6.1$ \\
$\{b,c\}$ & 3 & $+6.2$ \\
$\{a,c\}$ & 3 & $+7.0$ \\
$\{a,e\}$ & 2 & $+9.1$ \\
$\{c,e\}$ & 1 & $+11.2$ \\
\bottomrule
\end{tabular}

\smallskip
\raggedright
Sorted by $\widehat{\Delta L}$; \colorbox{cellAcc}{top candidate $\{a,b\}$} evaluated first.
\end{minipage}

\smallskip
\noindent\rule{\linewidth}{0.4pt}\smallskip


\noindent\textbf{Step~4: Clients evaluate $\{a,b\}$ — server verifies MDL gain and accepts}

\smallskip

\begin{minipage}[t]{0.46\linewidth}
\centering
\textit{Updated codetable $CT^*$ after inserting $\{a,b\}$}

\smallskip
\begin{tabular}{lccc}
\toprule
Itemset & $U^*$ & $L(c_X^*)$ \\
\midrule
\rowcolor{cellAcc}$\{a,b\}$ & 7 & 1.515 \\
$\{c\}$  & 4 & 2.322 \\
$\{d\}$  & 4 & 2.322 \\
$\{e\}$  & 3 & 2.737 \\
$\{a\}$  & 2 & 3.322 \\
\bottomrule
\end{tabular}

\medskip
\begin{tabular}{ll}
\toprule
& Cover (before $\to$ after) \\
\midrule
$t_1{=}\{a,b,c\}$: & $\covSg{a}+\covSg{b}+\covSg{c}$ $\to$ $\covAB{a,b}+\covSg{c}$ \\
$t_6{=}\{a,b,d\}$: & $\covSg{a}+\covSg{b}+\covSg{d}$ $\to$ $\covAB{a,b}+\covSg{d}$ \\
\bottomrule
\end{tabular}
\end{minipage}%
\hfill
\begin{minipage}[t]{0.50\linewidth}
\centering
\textit{MDL verification}

\medskip
\begin{tabular}{lll}
\toprule
 & $CT_0$ (before) & $CT^*$ (after) \\
\midrule
$L(D\mid CT)$    & 59.4\,bits & 44.0\,bits \\
$L(CT\mid D)$    & 24.4\,bits & 26.0\,bits \\
\midrule
$L_\text{total}$ & \textbf{83.9\,bits} & \textbf{70.1\,bits} \\
\bottomrule
\end{tabular}

\bigskip
\[
  L^* = 70.1 < L_0 = 83.9 \;\Rightarrow\;
  \boxed{\checkmark\;\textbf{Accept }\{a,b\};\;\text{broadcast and update}}
\]
\end{minipage}

\caption{Illustration of a complete \textsc{FedSLIM} iteration on a toy two-client federated database over items $\{a,b,c,d,e\}$.}
\label{fig:example}
\end{figure*}

%% file: figures/flow.tex
\begin{tikzpicture}[
    x=1cm, y=0.72cm,
    lifeline/.style={gray!60, dashed},
    entity/.style={font=\bfseries, align=center},
    msg/.style={-Latex, thick},
    selfmsg/.style={-Latex, thick},
    noteL/.style={draw, rounded corners, fill=gray!10, align=right, inner sep=4pt, font=\normalsize},
    noteR/.style={draw, rounded corners, fill=gray!10, align=left, inner sep=4pt, font=\normalsize},
    paneltitle/.style={
        font=\bfseries\fontsize{18}{20}\selectfont,
        text=black
    },
    every node/.style={font=\normalsize}
]


\def\xSOne{4}
\def\xCOne{11}

\def\xSTwo{-1}
\def\xCTwo{6}
\def\xPTwo{13}

\def\yTitleA{2}
\def\xTitleA{10}
\def\xTitleB{9}

\def\yTopA{-1}

\def\xLeftNote{-1.2}      
\def\xRightNote{4.5}      

\def\loopPadLeft{6.5}
\def\loopPadRight{7}

\def\selfLoopLooseness{4}

\def\rowGap{1.0}
\def\bigGap{1.6}
\def\phaseGap{1.4}
\def\selfDropRight{1}
\def\selfDropLeft{1}
\def\selfDropLeftTall{1}

\def\panelGap{4}


\newcommand{\setcursor}[2]{%
    \expandafter\xdef\csname y#1\endcsname{#2}%
}

\newcommand{\gety}[1]{%
    \csname y#1\endcsname
}

\newcommand{\step}[1]{%
    \pgfmathsetmacro{\tempY}{\gety{#1}-\rowGap}%
    \expandafter\xdef\csname y#1\endcsname{\tempY}%
}

\newcommand{\bigstep}[1]{%
    \pgfmathsetmacro{\tempY}{\gety{#1}-\bigGap}%
    \expandafter\xdef\csname y#1\endcsname{\tempY}%
}

\newcommand{\phasestep}[1]{%
    \pgfmathsetmacro{\tempY}{\gety{#1}-\phaseGap}%
    \expandafter\xdef\csname y#1\endcsname{\tempY}%
}


\newcommand{\msgat}[4]{%
    \draw[msg] (#2,\gety{#1}) -- (#3,\gety{#1})
        node[midway, above, sloped] {#4};
}

\newcommand{\selfright}[3]{%
    \pgfmathsetmacro{\tempY}{\gety{#1}-\selfDropRight}
    \draw[selfmsg] (#2,\gety{#1}) to[out=0,in=0,looseness=\selfLoopLooseness]
        node[right=0.2cm] {#3} (#2,\tempY);
}

\newcommand{\selfleft}[3]{%
    \pgfmathsetmacro{\tempY}{\gety{#1}-\selfDropLeft}
    \draw[selfmsg] (#2,\gety{#1}) to[out=180,in=180,looseness=\selfLoopLooseness]
        node[left=0.2cm, align=right] {#3} (#2,\tempY);
}

\newcommand{\selflefttall}[3]{%
    \pgfmathsetmacro{\tempY}{\gety{#1}-\selfDropLeftTall}
    \draw[selfmsg] (#2,\gety{#1}) to[out=180,in=180,looseness=\selfLoopLooseness]
        node[left=0.2cm, align=right] {#3} (#2,\tempY);
}

\newcommand{\phasenoteleft}[4]{%
    \node[noteL, anchor=west]
    at ({\xLeftNote + #3},{\gety{#1} + #4})
    {\textbf{#2}};
}

\newcommand{\phasenoteright}[4]{%
    \node[noteR, anchor=west]
    at ({\xRightNote + #3},{\gety{#1} + #4})
    {\textbf{#2}};
}

\newcommand{\loopboxA}[2]{%
    \draw[densely dotted, rounded corners]
        (\xSOne-8,#1)
        rectangle
        (\xCOne+6.3,#2);
}
\newcommand{\loopboxB}[2]{%
    \draw[densely dotted, rounded corners]
        (\xSTwo-\loopPadLeft,#1)
        rectangle
        (\xPTwo+\loopPadRight,#2);
}


\node[
    paneltitle,
    anchor=east,
    align=right,
    text width=10cm
] at (\xTitleA,\yTitleA) {\\\textsc{(a) FedSLIM-SA}};

\node[entity] at (\xSOne,0) {Server};
\node[entity] at (\xCOne,0) {All Clients};

\setcursor{A}{-0.8}

\phasenoteleft{A}{Registration}{-2.5}{-0.8}

\step{A}
\msgat{A}{\xCOne}{\xSOne}{Register ECDH keys}

\step{A}
\msgat{A}{\xSOne}{\xCOne}{Broadcast full public-key directory}

\selfright{A}{\xCOne}{\parbox{3cm}{\centering Derive pairwise seeds}}

\phasestep{A}
\phasenoteleft{A}{Initialization}{-2.5}{-0.8}

\step{A}
\msgat{A}{\xSOne}{\xCOne}{R1: Request ST index}

\step{A}
\msgat{A}{\xCOne}{\xSOne}{R1: Local structure (no counts)}

\selfleft{A}{\xSOne}{\parbox{3cm}{\centering Build global index\\(union + ordering)}}

\step{A}
\msgat{A}{\xSOne}{\xCOne}{R2: Broadcast global index}

\selfright{A}{\xCOne}{\parbox{3cm}{\centering Pad usages, \\apply SecAgg mask}}

\step{A}
\msgat{A}{\xCOne}{\xSOne}{R2: Masked aligned usage vectors}

\selfleft{A}{\xSOne}{\parbox{4cm}{\centering Secure aggregation,\\global CT initialization}}

\bigstep{A}

\step{A}
\pgfmathsetmacro{\outerLoopAStart}{\gety{A}+0.7}

\phasenoteleft{A}{Candidate Search}{-2.5}{-0.8}

\step{A}
\msgat{A}{\xSOne}{\xCOne}{Request candidate structures}

\selfright{A}{\xCOne}{\parbox{3.5cm}{\centering Run centralized SLIM \\candidates generation \\algorithm}}

\step{A}
\msgat{A}{\xCOne}{\xSOne}{Return candidate itemsets (XY, X, Y)}

\selfleft{A}{\xSOne}{\parbox{3.5cm}{\centering Deduplicate to \\global candidate index}}

\step{A}
\msgat{A}{\xSOne}{\xCOne}{Broadcast candidate index}

\selfright{A}{\xCOne}{\parbox{3.5cm}{\centering Pad $(\#\mathrm{cands} \times 3)$table, \\apply secAgg mask}}

\step{A}
\msgat{A}{\xCOne}{\xSOne}{Masked candidate statistics}

\selfleft{A}{\xSOne}{\parbox{5cm}{\centering Sum to aggregate stats,\\rank by estimated MDL gain}}

\bigstep{A}
\pgfmathsetmacro{\loopAStart}{\gety{A}-0.6}
\node[
    font=\normalsize\bfseries,
    anchor=south west,
    fill=white,
    inner sep=1.5pt
] at (\xSOne-8,\loopAStart) {Candidate Evaluation loop};
\step{A}

\selfleft{A}{\xSOne}{\parbox{2.8cm}{\centering Build $I_{\mathrm{eval}} = I^* \cup \{XY\}$}}

\step{A}
\msgat{A}{\xSOne}{\xCOne}{Evaluate candidate $XY$ + $I_{\mathrm{eval}}$}

\selfright{A}{\xCOne}{\parbox{5cm}{\centering If local: Run centralised SLIM candidate evaluation \\Else: keep current usage\\pad to $I_{\mathrm{eval}}$, mask}}

\phasenoteleft{A}{Update}{-2.5}{-0.8}

\step{A}
\msgat{A}{\xCOne}{\xSOne}{Masked post-insertion usages}

\selfleft{A}{\xSOne}{\parbox{5cm}{\centering Compute exact MDL gain\\Accept/reject candidate \\ Update global CT accordingly}}


\step{A}
\msgat{A}{\xSOne}{\xCOne}{If accepted: notify update}
\selfright{A}{\xCOne}{\parbox{3.5cm}{\centering If local: updated CT}}


\pgfmathsetmacro{\yAEnd}{\gety{A}-1.2}

\loopboxA{\loopAStart}{\yAEnd}

\draw[densely dashed, rounded corners]
    (\xSOne-8.3,\outerLoopAStart)
    rectangle
    (\xCOne+6.6,\yAEnd);

\node[
    font=\normalsize\bfseries,
    anchor=south west,
    fill=white,
    inner sep=1.5pt
] at (\xSOne-8.2,\outerLoopAStart) {SLIM main loop};

\draw[lifeline] (\xSOne,\yTopA) -- (\xSOne,\yAEnd);
\draw[lifeline] (\xCOne,\yTopA) -- (\xCOne,\yAEnd);

\draw[
    draw=black!70,
line width=0.8pt,
rounded corners=3pt,
    rounded corners=4pt
]
(\xSOne-9,\yTitleA+1)
rectangle
(\xCOne+7,\yAEnd-0.5);

%
%
\pgfmathsetmacro{\yBZero}{\yAEnd - \panelGap}
\pgfmathsetmacro{\yTitleB}{\yBZero + 1.5}
\pgfmathsetmacro{\yTopB}{\yBZero - 1}
\pgfmathsetmacro{\yBStart}{\yBZero - 0.8}


\node[
    paneltitle,
    anchor=east,
    align=right,
    text width=10cm
] at (\xTitleB,\yTitleB) {\\\textsc{(b) FedSLIM-SO}};

\node[entity] at (\xSTwo,\yBZero) {Server};
\node[entity] at (\xCTwo,\yBZero) {All Clients};
\node[entity] at (\xPTwo,\yBZero) {Participating Clients};

\setcursor{B}{\yBStart}

\phasenoteright{B}{Initialization}{13.7}{-0.8}

\step{B}
\msgat{B}{\xCTwo}{\xSTwo}{ST usages}

\selflefttall{B}{\xSTwo}{\parbox{5cm}{\centering Usages aggregation, \\Global CT initialization}}

\step{B}
\phasestep{B}
\pgfmathsetmacro{\outerLoopBStart}{\gety{B}+0.4}
\phasenoteright{B}{Candidate Search}{13.7}{-0.8}

\step{B}
\msgat{B}{\xSTwo}{\xCTwo}{Request candidates}

\selfright{B}{\xCTwo}{\parbox{3.5cm}{\centering Run centralized SLIM \\candidates generation}}

\step{B}
\msgat{B}{\xCTwo}{\xSTwo}{Candidates tuples}

\selflefttall{B}{\xSTwo}{\parbox{4.5cm}{\centering Merge by $XY$, sum stats\\Track participants \\Rank by MDL estimated gain}}

\bigstep{B}
\step{B}

\pgfmathsetmacro{\loopBStart}{\gety{B}-0.4}
\node[
    font=\normalsize\bfseries,
    anchor=south west,
    fill=white,
    inner sep=1.5pt
] at (\xSTwo-6.5,\loopBStart) {Candidate Evaluation loop};

\bigstep{B}

\msgat{B}{\xSTwo}{\xPTwo}{Evaluate candidate $XY$}

\selfright{B}{\xPTwo}{\parbox{4cm}{\centering Run centralised SLIM \\candidate evaluation}}

\step{B}
\msgat{B}{\xPTwo}{\xSTwo}{Post-insertion usages}

\selflefttall{B}{\xSTwo}{\parbox{5cm}{\centering Combine with cached\\non-participant states,\\Compute exact MDL gain,\\Accept/reject \& Update global CT and cache accordingly}}

\step{B}
\phasenoteright{B}{Update}{13.7}{0}
\msgat{B}{\xSTwo}{\xPTwo}{If accepted: notify update}

\selfright{B}{\xPTwo}{\parbox{3.5cm}{\centering updated local CT}}



\pgfmathsetmacro{\yBEnd}{\gety{B}-1.2}

\loopboxB{\loopBStart}{\yBEnd}

\draw[densely dashed, rounded corners]
    (\xSTwo-7,\outerLoopBStart)
    rectangle
    (\xPTwo+8.7,\yBEnd);

\node[
    font=\normalsize\bfseries,
    anchor=south west,
    fill=white,
    inner sep=1.5pt
] at (\xSTwo-7,\outerLoopBStart) {SLIM main loop};

\draw[lifeline] (\xSTwo,\yTopB) -- (\xSTwo,\yBEnd);
\draw[lifeline] (\xCTwo,\yTopB) -- (\xCTwo,\yBEnd);
\draw[lifeline] (\xPTwo,\yTopB) -- (\xPTwo,\yBEnd);

\draw[
    draw=black!70,
line width=0.8pt,
rounded corners=3pt,
    rounded corners=4pt
]
(\xSTwo-7.5,\yTitleB+1)
rectangle
(\xPTwo+9,\yBEnd-0.5);

\end{tikzpicture}